\documentclass{article}

\PassOptionsToPackage{numbers,compress}{natbib}
\usepackage[preprint]{neurips_2026}

\usepackage[utf8]{inputenc}
\usepackage[T1]{fontenc}
\usepackage{hyperref}
\usepackage{url}
\usepackage{booktabs}
\usepackage{amsfonts}
\usepackage{amsmath}
\usepackage{amssymb}
\usepackage{nicefrac}
\usepackage{microtype}
\usepackage[table]{xcolor}
\usepackage{graphicx}
\usepackage{array}
\usepackage{ragged2e}
\usepackage{makecell}
\usepackage{enumitem}
\usepackage{wrapfig}

\newcolumntype{L}[1]{>{\RaggedRight\arraybackslash}p{#1}}

\providecommand{\Description}[1]{}

\makeatletter
\newcommand{\sidecaptiontable}[2]{%
  \refstepcounter{table}\label{#1}%
  \addcontentsline{lot}{table}{\protect\numberline{\thetable}{#2}}%
  \@makecaption{\tablename~\thetable}{#2}%
}
\makeatother

\title{Memory of Memory}

\makeatletter
\if@anonymous
  \newcommand{\CorrespondingAuthor}{}
\else
  \newcommand{\CorrespondingAuthor}{\thanks{Corresponding author.}}
\fi
\makeatother

\author{%
  Bowen Qin \\
  National University of Singapore \\
  \texttt{qin.bowen@u.nus.edu} \\
  \And
  Yao Lu\CorrespondingAuthor{} \\
  National University of Singapore \\
  \texttt{yao@nus.edu.sg} \\
}

\begin{document}

\maketitle

\begin{abstract}
For a long-horizon LLM agent, the memory question is not what was once recorded but what \emph{currently holds}. Most designs answer it only indirectly: every interaction is stored, and the present is reconstructed at query time by retrieving and reconciling records, so stale values re-enter and the same conflicts are re-litigated. Committing the current value at write time avoids this, but existing write-time (CRUD) memories overwrite, so a wrong update is unrecoverable and prior state is lost. We take the missing combination---\emph{commit on arrival while retaining what is displaced}---and formalize it as \textsc{Memory of Memory} (MoM): memory tracks not only content but the provenance, status, and history of its own entries. We instantiate MoM as \textsc{Provenant Memory} (P-Mem), a typed provenance graph whose \emph{active frontier} exposes one current value per resolved key while displaced values are retained as provenance; typed operations decide whether a new observation supports, supersedes, contests, rejects, revokes, or resolves an existing value.
P-Mem's decisive gain is validity rather than accuracy: its turn-level read matches the strongest retrieval memory in accuracy at $\sim$4$\times$ fewer read tokens---a retrieval-granularity effect---while graph-guided turn pruning cuts the knowledge-update stale-answer rate (19.4\%$\rightarrow$10.9\%); on revision chains it stays at 100\% where query-time reading collapses to 25\%, and, because displaced values are retained rather than overwritten, it recovers committed errors a CRUD memory cannot (100\% vs.\ 0\%).
\end{abstract}

\section{Introduction}
\begin{figure}[t]
  \centering
  \includegraphics[width=0.9\linewidth]{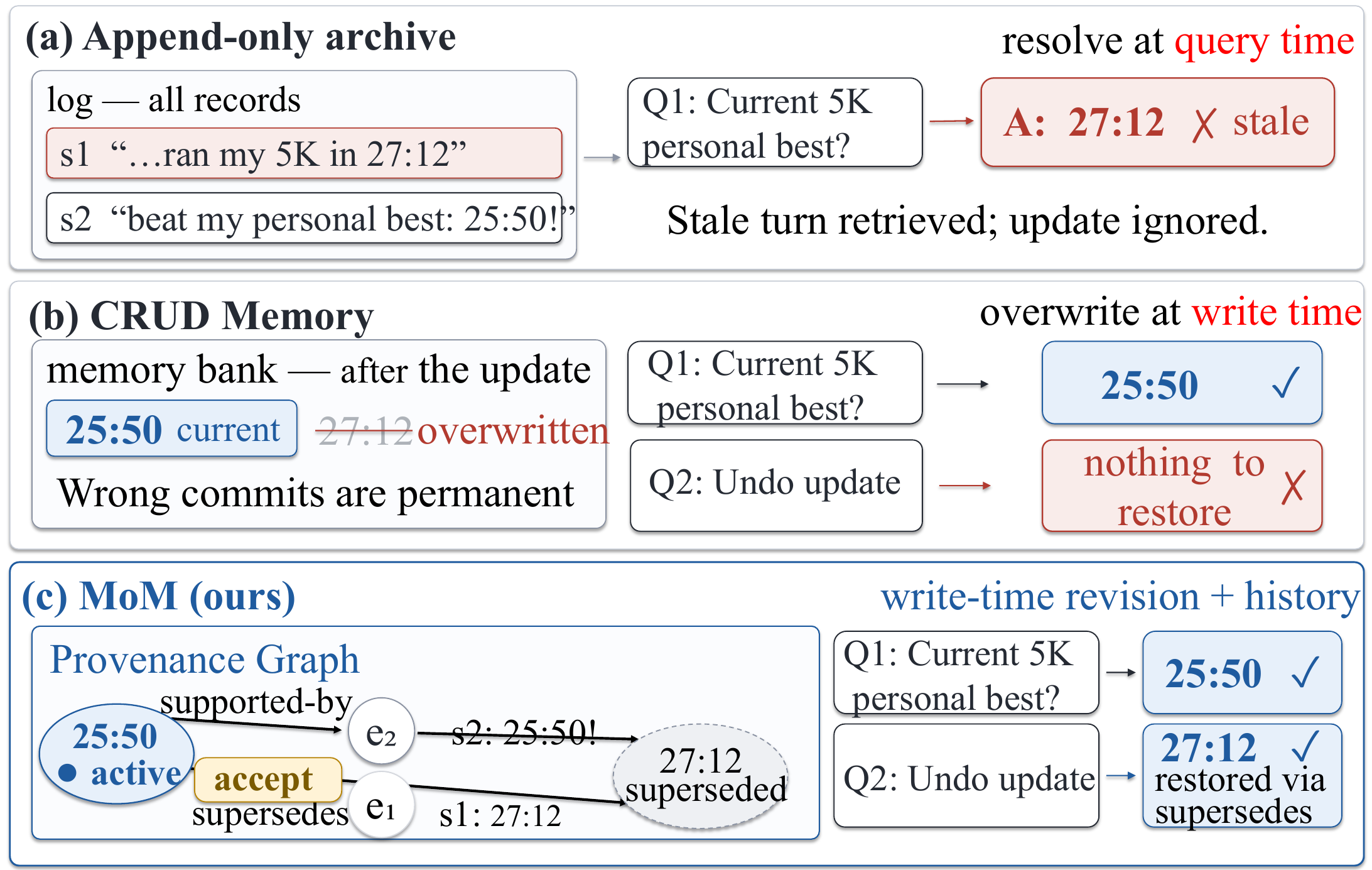}
    \caption{One knowledge update across three memory designs. (a) The archive returns the stale value; (b) CRUD memory cannot undo a mistaken update; (c) MoM reads the current value and restores it on retraction.} 
  \Description{Three side-by-side panels compare an append-only archive, a CRUD memory, and MoM on one revised fact. The archive returns the stale value; the CRUD store answers the current value but cannot undo a mistaken update; MoM answers the current value and restores the prior value after a retraction.}
  \label{fig:provenance-case}
\end{figure}

An LLM agent operating over a long horizon must answer what \emph{currently holds}, not merely recall what was once said. Consider an assistant (Figure~\ref{fig:provenance-case}) told that a user's 5K personal best is 27:12, then months later 25:50. Asked ``what's my 5K PB?'', a retrieval-based memory typically returns \emph{27:12}: the earlier turn states the value as a plain fact and matches the query, while the newer value appears only in passing (``beat my PB of 25:50''). It faithfully reports a stale value. The difficulty is not storage but \emph{evidential status}---which value is current, what it superseded, and why it is still trusted.

Treat memory as a notebook and ask what it does when a fact changes. Two choices are decisive (Table~\ref{tab:design-space}): whether it \emph{settles the current value on arrival} or defers to query time, and whether it \emph{retains or erases} the value it replaces. Existing systems take one of two combinations. \emph{Archival} memory---long-context, RAG, note-style, value-aware retrieval~\cite{zhang2026memrl}---defers and retains: all records stay active and the present is reconstructed per query, so nothing is marked stale and the outdated ``27:12'' keeps resurfacing, the same conflict re-adjudicated, sometimes inconsistently. \emph{CRUD} memory with update and delete, e.g.\ Mem0~\cite{chhikara2025mem0}, settles on arrival but overwrites: a wrong update is unrecoverable, a genuine conflict forced to one answer, prior state lost.

\begin{table}[t]
\centering
\caption{\textbf{Memory designs along two axes:} when the current value is settled (on arrival vs.\ query time) and whether the displaced value is retained or erased. MoM commits on arrival yet retains what it displaces as typed provenance; bitemporal KGs share that corner but arbitrate by timestamp, leaving conflict implicit and commits irreversible.}
\label{tab:design-space}
\footnotesize
\setlength{\tabcolsep}{5pt}
\renewcommand{\arraystretch}{1.25}
\begin{tabular}{@{}L{0.31\linewidth}>{\centering\arraybackslash}p{0.11\linewidth}L{0.17\linewidth}L{0.29\linewidth}@{}}
\toprule
\textbf{Design family (systems)} &
\textbf{\makecell{Current value\\settled}} &
\textbf{Displaced value} &
\textbf{Characteristic failure} \\
\midrule
Archival: long-context, RAG / note-style, value-aware retrieval (MemRL), tiering (MemGPT / MemoryOS) &
query time &
retained, not demoted &
staleness re-entry; conflicts re-litigated, sometimes inconsistently \\
\addlinespace
CRUD memory (Mem0-style banks) &
on arrival &
erased (overwritten / deleted) &
mis-updates irreversible; history unanswerable \\
\addlinespace
Bitemporal KG (Zep / Graphiti) &
on arrival &
retained (valid-time intervals) &
timestamp-arbitrated; no open conflict, no reversal \\
\addlinespace
\textbf{MoM (ours)} &
on arrival &
\textbf{retained as provenance} &
residual: wrong until revisited (\S\ref{sec:revision-results}) \\
\bottomrule
\end{tabular}
\end{table}

\vspace{0.02in}\noindent\textbf{Memory of Memory (MoM)}. What we want is the missing combination: \emph{commit on arrival while retaining what is displaced}. Memory maintains a compact \emph{current view}---one value per key, updated at write time, so the agent reads the answer rather than reconstructing it per query---while every displaced value is kept with its \emph{provenance}: where it came from and what it replaced.

We instantiate MoM as \textsc{Provenant Memory} (P-Mem): each observation triggers one typed operation---\textsc{accept} (supersede the incumbent), \textsc{contest} (record an unresolved alternative), \textsc{reject} (mark the incumbent wrong), \textsc{revoke}, \textsc{resolve} (settle a standing conflict), or attach support with no change. Two messages naming different meeting rooms with no precedence cue invoke \textsc{contest}, keeping both alternatives rather than guessing until a later message triggers \textsc{resolve}. On the example, 25:50 becomes current and 27:12 is retained as \emph{superseded}: ``PB now?'' reads the current value, ``old PB?'' stays answerable, a mistaken update stays reversible. Each operation asserts a \emph{relation}---one value supersedes another, competes as an alternative, or draws support from an observation---and it is these relations, not the retained values alone, that let a mistaken commit be reversed, a conflict held open, and a value justified.

\vspace{0.02in}\noindent\textbf{Provenance graph.} Recording and traversing such relations is what \emph{data provenance}~\citep{buneman2001why,moreau2013prov} and \emph{temporal databases}~\citep{jensen1999temporal} were built for, so P-Mem borrows their central object and casts each fact's history as a \emph{provenance graph}: two kinds of node---the observations received and the values derived from them---joined by typed edges, where a value \emph{supersedes} an earlier one, stands as an \emph{alternative} to a competitor, or is \emph{supported by} an observation. Each fact's current value is the graph's \emph{active frontier}, the readout the agent sees; everything it displaces stays reachable behind it, so history, justification, and rollback are backward traversals of these edges---the current value cheap to read, the store correctable and auditable rather than reconciled afresh per query. Why the edges, not retention alone, matter: memory records a client budget of \$30k, reads ``raised to \$50k'' and makes \$50k current; a week later the user retracts \$50k as an error. Reinstating \$30k needs the recorded edge that \$50k \emph{superseded} it---which a flat list cannot supply: CRUD has discarded \$30k, and a flat archive retains it but cannot identify it as the value \$50k replaced.

\vspace{0.02in}\noindent\textbf{Building and reading the graph.} Classical provenance and temporal databases assume a fixed schema, a correct transactional writer, and a reader free to scan the whole store; an agent setting violates all three, so P-Mem meets each with one mechanism. {(i) Keys are language-inferred, not schema-fixed}, so one attribute surfaces in many forms (``5K PB,'' ``personal best over 5\,km''): \emph{soft key identity}---an embedding match confirmed by a light check---routes each update to the intended cell at write time rather than forking a stale duplicate. {(ii) The writer is a fallible LLM and currency cannot be re-derived at read time}, so \emph{write-time supersession by arrival order} settles the frontier as observations arrive, sparing the reader from reassembling a revision chain from similarity-ranked text---the failure that compounds as chains grow (\S\ref{sec:revision-results}). {(iii) The reader is context-budgeted}, so \emph{deterministic materialization with disclosure modes} renders each cell current-value-first and walks provenance edges only on demand, keeping a read token-bounded with justification one hop away (\S\ref{sec:token-efficiency}). Retention ties these together: nothing is deleted, so a mis-commit persists as a demoted node on its \emph{supersedes} edge, reinstated by a later retraction instead of failing (\S\ref{sec:retraction}).

These mechanisms yield measurable validity gains across three benchmarks. On LongMemEval-s, P-Mem matches the strongest retrieval baseline in accuracy (52.6\% vs.\ 50.8\% for a graph-RAG memory, within noise) at about 4$\times$ fewer context tokens; its distinctive gain is validity---graph-guided turn pruning cuts the knowledge-update stale rate from 19.4\% to 10.9\%. On ALFWorld-Revision chains, write-time commit holds 100\% as revisions accumulate, where a query-time read that must reassemble the chain falls to 25\%; a retraction recovers the displaced value (100\% vs.\ 0\% for CRUD), and contest/resolve settles a genuine conflict once rather than per query. On MemoryAgentBench FactConsolidation, at matched single-read cost write-time consolidation improves multi-hop accuracy 1.4--2.9$\times$ over the order-tagged flat store---iterative disclosure lifts it further---while single-hop stale rates stay at 3--7\% as history grows 40-fold.

\vspace{0.02in}\noindent\textbf{Contributions} of this paper can be summarized as follows:
\begin{itemize}[leftmargin=*,itemsep=1pt,topsep=2pt]
\item We recast agent memory as maintaining current state at write time, not reconstructing it per query, and name the missing design point---commit on arrival yet retain what is displaced (Table~\ref{tab:design-space}).
\item We introduce \textsc{Provenant Memory} (P-Mem), a typed provenance graph whose \emph{active frontier} gives one current value per resolved key, displaced values retained, via typed operations and three mechanisms: soft key identity, write-time supersession, disclosure-mode materialization.
\item Across three benchmarks (with our ALFWorld-Revision) validity, not accuracy, is decisive: it matches the strongest retriever at $\sim$4$\times$ fewer tokens, pruning halves the knowledge-update stale rate, chains hold 100\% vs.\ 25\%, and committed errors recover where CRUD loses them.
\end{itemize}


\section{Related Work}
\label{sec:related-work}

\vspace{0.02in}\noindent\textbf{Agent memory as an online lifecycle.}
Beyond treating memory as context selection over stored history, recent surveys and benchmarks frame it as an online lifecycle---information written, consolidated, retrieved, updated, and forgotten as the agent interacts~\citep{zhang2024memorysurvey,hu2025memoryagentbench,wei2025evomemory,zhang2026neuromem}. Long-term evaluations (LoCoMo, LongMemEval, MemoryArena) reflect the shift: success turns on updates, temporal dependencies, and downstream action, not recall alone~\citep{maharana2024locomo,wu2025longmemeval,he2026memoryarena}. MoM presses the two questions left open (Table~\ref{tab:design-space}): \emph{when} a fact's current value is settled---on arrival or at query time---and what becomes of the value it displaces.

\vspace{0.02in}\noindent\textbf{Where existing systems fall in the design space.}
Under these axes, most memory systems occupy the archival or CRUD corners of Table~\ref{tab:design-space}. Tiered systems~\citep{packer2023memgpt,kang2025memoryos} manage \emph{placement}---what stays in context under a size limit---not which value holds; so do structural, agentic-linking, hierarchical, and preference-aware memories~\citep{zeng2024structuralmemory,xu2025amem,sun2025hierarchicalmemory,sun2025pamu} and graph-RAG archives (HippoRAG-v2)~\citep{gutierrez2025hipporag2}, which organize evidence richly but resolve conflicts per query. Write-side systems~\citep{chhikara2025mem0,zhang2026memskill,zhang2026uma} move consolidation to write time---the right direction---but overwrite, so a wrong commitment is unrecoverable from memory. Closest to us, bitemporal knowledge graphs such as Zep/Graphiti~\citep{rasmussen2025zep} commit on arrival and retain superseded edges; general temporal and event-sourced stores can in principle model corrections and rollback, but as applied to agent memory they arbitrate by timestamp and expose neither an unresolved conflict as first-class state nor a reopened commit (\S\ref{sec:retraction}, \S\ref{sec:conflict}). Our novelty is therefore not retention itself but the agent-specific combination: language-inferred soft keys, fallible-LLM operation selection, first-class conflict state, and compact provenance-aware disclosure read from the active frontier.

\vspace{0.02in}\noindent\textbf{Retrieval and stale evidence in agent memory.}
Utility-aware retrieval such as MemRL learns which experiences help future decisions, and retrieval choices can even dominate write-time construction~\citep{zhang2026memrl,yuan2026diagnosing}; yet these improve \emph{which} record is selected without committing a state. Stale memories left active also propagate errors~\citep{xiong2025memorymanagement}---superseded records should be demoted to provenance, not left co-active. The database fix---retain, but mark superseded---breaks three assumptions in the agent setting: keys are extracted from language, not a schema; the writer is a fallible LLM, not a correct transaction; and the reader is context-budgeted, not free to scan the store. P-Mem answers with soft key identity (Appendix~\ref{app:mabcr-extra}), write-time supersession by arrival order (\S\ref{sec:revision-results}), and disclosure-mode materialization (\S\ref{sec:token-efficiency}), while retention keeps a mistaken commit recoverable (\S\ref{sec:retraction}). Better retrieval selects \emph{which record is relevant}; MoM commits \emph{which value holds}, and \S\ref{sec:revision-results} measures its cost.

\section{Method}
\label{sec:method}

\subsection{Memory-Augmented Agent Policy}
\label{sec:problem-setup}

A memory-augmented agent adapts across a task stream through an external
memory state while keeping its base policy fixed. Let \(q_t\) be the task at
step \(t\) and \(z_{t-1}\) the memory accumulated before it. The memory
system produces an agent-facing context
\begin{equation}
\label{eq:generic-read}
  \hat{c}_{t-1}(q_t)
  =
  \operatorname{Read}(z_{t-1},q_t),
  \qquad
  a_t
  \sim
  p_{\mathrm{LLM}}
  \left(
    \cdot
    \mid
    q_t,\hat{c}_{t-1}(q_t)
  \right).
\end{equation}
The base LLM is fixed; only the context it receives from memory changes.
\begin{figure}
    \centering
    \includegraphics[width=0.9\linewidth]{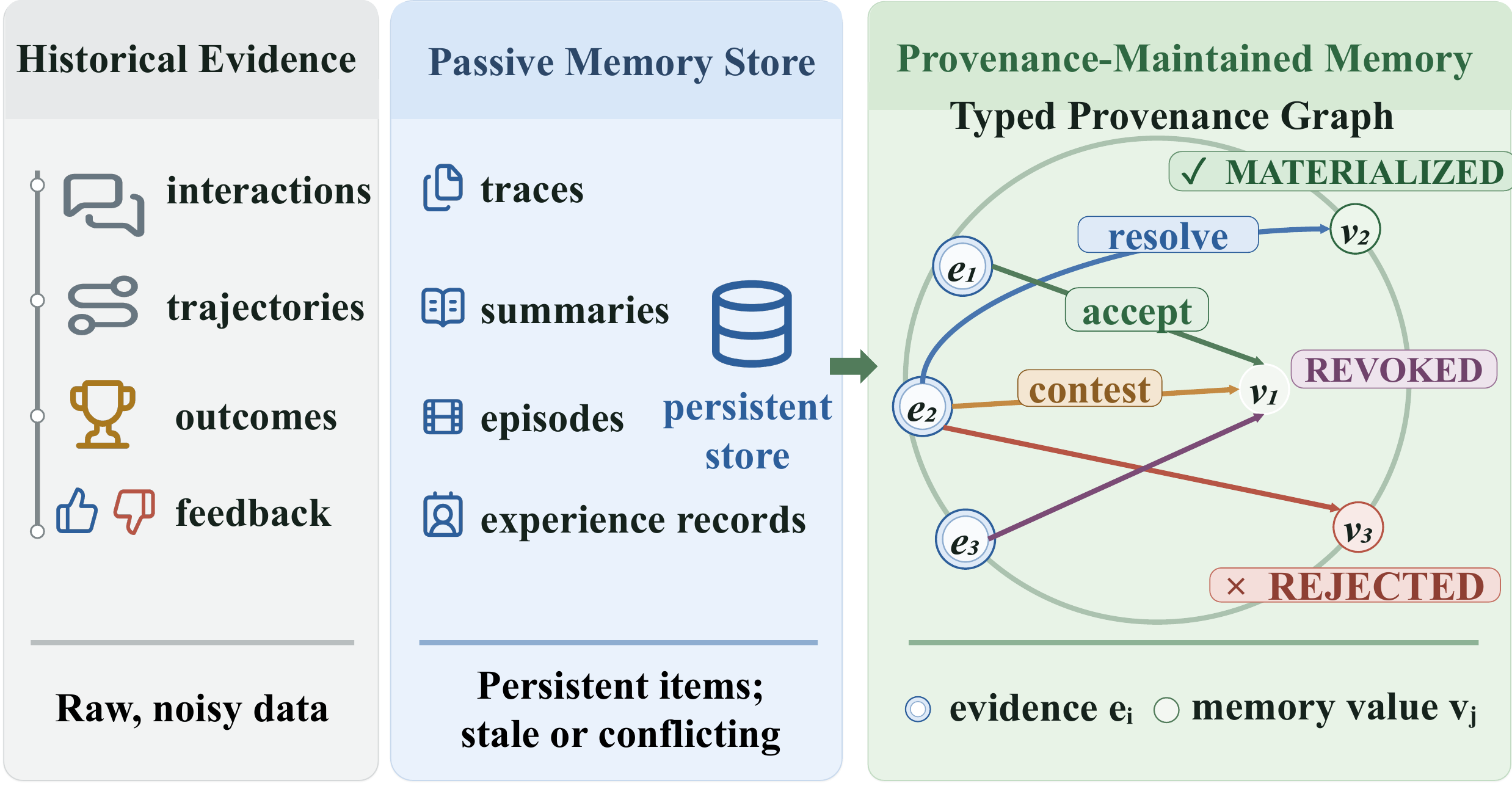}
    \caption{P-Mem overview. A typed temporal provenance graph turns passive interaction records into an evidence-grounded active frontier of currently valid values, while preserving displaced values as provenance.}
    \label{fig:memory-layers}
\end{figure}

After execution, the state is updated from the completed record
\begin{equation}
\label{eq:generic-memory-update}
  z_t
  =
  \operatorname{Update}(z_{t-1},x_t),
  \qquad
  x_t=(q_t,\tau_t,y_t),
\end{equation}
with \(\tau_t\) the execution trajectory and \(y_t\) the observed outcome or feedback.

Existing agents instantiate \(z_{t-1}\) as historical evidence and implement \(\mathrm{Read}\) by retrieving items relevant to the task. This beats conditioning on raw trajectories but leaves one question open: \textbf{relevance does not imply validity}---a relevant item may be outdated, contradicted by later evidence, or superseded. P-Mem organizes items and their evidence into the typed temporal provenance graph of Figure~\ref{fig:memory-layers} (developed next), maintaining an \emph{active frontier} of currently valid values: a relevant memory is withheld when newer evidence contradicts, revokes, or supersedes it, turning long-term memory from a passive store into an actively maintained decision state.

\subsection{The provenance graph}
\label{sec:pmem}

P-Mem represents the complete memory record as a typed temporal provenance graph,
\begin{equation}
\label{eq:provenance-graph}
  z_t := G_t = (V_t,\mathcal{R}_t),
\end{equation}
with four node kinds and typed edges. \emph{Evidence nodes} hold a raw observation with source and timestamp; \emph{version nodes} a (key, value) pair with a status and validity interval; \emph{operation nodes} an applied update and its time; \emph{tombstone nodes} mark revocations. Edges are \emph{supported-by} (version to its evidence), \emph{supersedes} and \emph{alternative-of} (version to the predecessor it replaces or the sibling it competes with), and \emph{revokes} (tombstone to the version it retracts).

A single fact is thus a \emph{subgraph} (Figure~\ref{fig:memory-layers}): the versions a key has held, the evidence for each, and the operations that moved authority between them. The graph is append-only---nothing admitted is deleted, only retyped or relinked---so what currently holds and everything it displaced both remain, distinguished by status and edge type.

\vspace{0.02in}\noindent\textbf{Typed operations.}
A fact enters the graph through one of a fixed set of operations relating it to the key's current head,
\begin{equation}
\label{eq:op-types}
  \mathrm{type}(\omega) \in \{\textsc{accept}, \textsc{contest}, \textsc{reject}, \textsc{revoke}, \textsc{resolve}, \textsc{noop}\},
\end{equation}
each writing typed edges and setting version status:
\begin{itemize}[leftmargin=*,itemsep=1pt,topsep=2pt]
  \item \textsc{accept}: the fact becomes the head (\emph{active}); the incumbent is retyped \emph{superseded} and kept.
  \item \textsc{contest}: the fact attaches as an \emph{alternative}; the head becomes \emph{contested}, holding provisional authority until settled.
  \item \textsc{reject}: the fact becomes the head; the incumbent is retyped \emph{contradicted}---recorded wrong, not merely outdated.
  \item \textsc{revoke}: the head is retracted (\emph{revoked}) and a tombstone written; the cell may become empty.
  \item \textsc{resolve}: one alternative of a contested cell is promoted to head, the others retyped \emph{contradicted}.
  \item \textsc{noop}: no version or status change; adds a \emph{supported-by} edge and increments the head's support count.
\end{itemize}
While a cell is contested, the frontier still exposes the incumbent, flagged as such, with alternatives reachable through the \texttt{alternatives} disclosure mode (\S\ref{sec:reading}).

Two passes follow: \S\ref{sec:memory-updates} \emph{writes} the graph, applying these operations as observations arrive, and \S\ref{sec:reading} \emph{reads} it, exposing current values through a fixed materialization.

\subsection{Writing: validity-aware updates}
\label{sec:memory-updates}

P-Mem folds the completed record \(x_t\) into the graph in two stages: extraction turns language into keyed facts, then a typed operation places each relative to the current state.

\vspace{0.02in}\noindent\textbf{Extraction and soft identity.}
A frozen extractor maps the record to a set of typed facts, each a value with a candidate key naming a subject and attribute (e.g.,\ \texttt{personal\_best\_5k\_time}):
\begin{equation}
\label{eq:extract}
  \{r_{t,j}\}_{j=1}^{n_t} = \operatorname{Extract}(x_t).
\end{equation}
Because keys are language-inferred rather than schema-fixed, one attribute can surface under different names, and extraction alone fragments equivalent keys---on the 6k MemoryAgentBench stream, 346 cells for 239 template keys---so without repair a later mention forks a duplicate that leaves a stale value active. \emph{Soft identity} reconciles this at write time: the candidate key is embedded, its most similar registered key shortlisted when cosine reaches $0.72$, and a one-word same/different call shown only the two attribute names merges them on \textsc{same}. Both stages use keys alone---including the values collapses the merge (the verifier accepts 2 of 135 pairs vs.\ 37 of 133 on keys alone), since equal values recur across distinct keys. Deciding which cell an observation updates is thus an inference step absent from schema-based provenance.

\vspace{0.02in}\noindent\textbf{Applying operations.}
For each fact, a frozen annotator selects one typed operation (Eq.~\ref{eq:op-types}) relating it to the key's head, schema-constrained and validated by deterministic post-processing (\textsc{contest} needs an active incumbent, \textsc{resolve} a contested cell; a failing annotation falls back to \textsc{noop}). A record's operations apply to \(G_{t-1}\) in arrival order, and \(M_t\) re-materializes from the result.

Arrival order fixes the direction of supersession at write time, so the reader never reconstructs it at query time. Nothing is deleted---superseded, contradicted, and revoked versions stay in \(G_t\) as provenance---which makes a mistaken \textsc{accept} reversible by a later \textsc{revoke}/\textsc{reject} (\S\ref{sec:retraction}) and lets a conflict be held by \textsc{contest} and settled once by \textsc{resolve} (\S\ref{sec:conflict}), neither of which a CRUD writer supports. Table~\ref{tab:annotation-lowering} summarizes the lowering onto graph edges and statuses.

\begin{table}[t]
\centering
\caption{How each typed annotation (\S\ref{sec:pmem}) lowers onto version-status changes. \emph{Head after} is the cell's current value once applied; every other version is retained, retyped as listed.}
\label{tab:annotation-lowering}
\small
\setlength{\tabcolsep}{4pt}
\begin{tabular}{@{}lll@{}}
\toprule
\textbf{Annotation} & \textbf{Head after} & \textbf{Other version(s)} \\
\midrule
\textsc{accept}  & new value (active)            & superseded \\
\textsc{reject}  & new value (active)            & contradicted \\
\textsc{contest} & incumbent (contested)        & incoming kept as alternative \\
\textsc{revoke}  & predecessor, else empty       & revoked (tombstoned) \\
\textsc{resolve} & promoted alternative (active) & others contradicted \\
\textsc{noop}    & unchanged                     & unchanged \\
\bottomrule
\end{tabular}
\end{table}

\vspace{0.02in}\noindent\textbf{Patch interface.} In the benchmark write paths, facts reach the annotator through one patch form---a \textsc{revise}-shaped object with \texttt{op}, \texttt{key}, \texttt{old\_value}, \texttt{new\_value}, \texttt{status}, \texttt{supersedes}---which a deterministic controller classifies against the head: a \textsc{create} for a new key, else a supersession (\textsc{accept}) or revision (\textsc{reject}). A hard-commit flag forces application when a structured correction must not be dropped. The ALFWorld-Revision and MemoryAgentBench write paths both use it.

\vspace{0.02in}\noindent\textbf{Rollback along the supersedes chain.} Because nothing is deleted, a retraction is a graph walk, not a re-derivation. \textsc{revoke} closes the head's interval, writes a tombstone with a \emph{revokes} edge, and marks the head revoked; the new head is the most recent predecessor along the retracted version's \emph{supersedes} edge, whose interval reopens as active (the cell empties only if none exists). A \textsc{reject} with no replacement rolls back identically but marks the retracted head \emph{contradicted} rather than \emph{revoked}; a \textsc{reject} carrying a new value installs it as head instead of reopening the predecessor (Table~\ref{tab:annotation-lowering}). Consecutive retractions walk the chain one predecessor at a time, so $k$ retractions reinstate the value from before the $k$ most recent commits; the chain is acyclic because \emph{supersedes} points strictly backward. As the reinstated value is read from a retained node rather than reconstructed, rollback is deterministic---the mechanism behind the perfect error recovery of \S\ref{sec:retraction}.

\subsection{Reading: materialization and readout}
\label{sec:reading}
\label{sec:provenance-readout}

A fixed, query-independent materialization derives a structured state \(M_t = \operatorname{Materialize}(G_t)\), organized by key. For each key it selects the \emph{head}---the most recent open version with status \textsc{active} or \textsc{contested}---as the current value; a functional-key invariant demotes any other open version, so no key exposes two current values. The heads form the \emph{active frontier} the agent treats as current. Statuses range over \{active, contested, superseded, contradicted, revoked, historical\}---\emph{contested} being a challenged head with its alternatives, \emph{historical} any closed-interval version. Each cell retains its evidence, alternatives, and closed versions as provenance. Because materialization is deterministic and query-independent, the current value is committed once at write time, not re-derived per query---the property separating P-Mem from retrieval memories. A cell renders as one line, \texttt{key = value}, with support, alternatives, or history appended on demand.

Given \(q_t\), the readout turns this state into the agent context \(\hat{c}_{t-1}(q_t) = \operatorname{Read}_{\phi}(M_{t-1},q_t)\) without creating evidence or modifying \(G_{t-1}\); the pipeline is \(G_t \to M_t \to \hat{c}_t(q)\). A retriever forms a bounded set \(\mathcal{C}_t = \operatorname{Retrieve}_{L}(M_{t-1},q_t)\) by embedding each cell as a \texttt{key: value} string and ranking by cosine under a per-benchmark top-\(L\) cutoff, and a disclosure mode \(\delta_t\) exposes, with each current value, a subset \(P_{t,i} \subseteq \{\mathtt{support}, \mathtt{alternatives}, \mathtt{history}\}\). The interface admits a learned policy \(\pi_\phi(\delta_t \mid q_t, \mathcal{C}_t)\), but every experiment uses a deterministic default rendering each cell current-value-first at fixed depth, so the readout has no trained parameters; the retraction and conflict probes (\S\ref{sec:retraction}, \S\ref{sec:conflict}) disclose \(\mathtt{history}\) and \(\mathtt{alternatives}\) through it. It has two instantiations: a direct read of materialized cells (MemoryAgentBench, ALFWorld-Revision) and, on LongMemEval, a turn-level retrieval whose superseded turns are graph-pruned (Appendix~\ref{app:materialization}); both expose current values and withhold superseded ones.

\vspace{0.02in}\noindent\textbf{Graph-guided pruning.} The turn-level path enforces validity even over raw turns: when two retrieved turns are related by supersession, the earlier is dropped for the survivor, so a superseded statement cannot re-enter through similarity alone; arrival order recorded at write time fixes which is stale, and the reader never judges recency (Appendix~\ref{app:config-lme}). This is the only place a graph relation acts on a non-materialized read, and it separates P-Mem's stale rate from a plain top-$k$ retriever on the knowledge-update subset (\S\ref{sec:longmemeval-results}).

\vspace{0.02in}\noindent\textbf{Activated subset.} The data model carries more than the runs exercise---usage edges, support-count ranking, learned retrieval/retention scores, budget eviction---all inactive here, so every reported number uses deterministic retrieval and the fixed disclosure default (Appendix~\ref{app:implementation}). P-Mem is thus evaluated on its typed write operations and query-independent materialization, not a tuned reader.

\section{Experiments}
\label{sec:experiments}

\vspace{0.02in}\noindent\textbf{Research questions.}
Our evaluation isolates the properties that distinguish write-time commitment with retention (Table~\ref{tab:design-space}) from archival and CRUD memory:
\begin{itemize}[leftmargin=*]
\item \textbf{RQ1 (validity vs.\ accuracy).} On knowledge updates, (a) does graph-guided pruning of retrieved turns lower the stale-answer rate while keeping accuracy competitive with the strongest retrieval memories; and (b) does committing supersession at write time, answering from materialized state alone, do the same on naturally phrased histories?
\item \textbf{RQ2 (read cost).} Does reading the active frontier answer at lower context-token cost than retrieval or full-history reads?
\item \textbf{RQ3 (revision chains).} As revisions to one fact accumulate, does write-time commit stay current where a query-time read must reassemble arrival order?
\item \textbf{RQ4 (reversibility).} Does retention turn a committed error from a permanent failure into a transient one, recoverable once the error is retracted?
\item \textbf{RQ5 (conflict).} Can memory represent a genuine conflict instead of guessing, and settle it once when evidence arrives?
\item \textbf{RQ6 (scale and multi-hop).} As the history grows, does write-time consolidation preserve single-hop currency and multi-hop accuracy where flat, CRUD, and archival stores degrade?
\end{itemize}

\vspace{0.02in}\noindent\textbf{Datasets.}
We use three benchmarks, each stressing a different part of the update lifecycle.
\emph{LongMemEval}~\citep{wu2025longmemeval} poses questions over a chat history where a fact is stated and later revised. We use its 78 knowledge-update questions (the current value is asked after revision) in two haystack regimes---\emph{oracle} (evidence sessions only, $\sim$2 per question) and the full \emph{LongMemEval-s} ($\sim$48 sessions per question)---and the 500-question superset for the full-benchmark read (\S\ref{sec:longmemeval-results}).
\emph{ALFWorld-Revision} is our revision-pressure variant of ALFWorld~\citep{shridhar2021alfworld}, a text-based embodied environment for household tasks. Standard ALFWorld never invalidates a stored fact---object states change only through the agent's own actions and can be re-observed---so we add revisions: an object's stated location is changed by one or more updates and the agent must report its current location.
\emph{MemoryAgentBench FactConsolidation}~\citep{hu2025memoryagentbench} is the benchmark's conflict-resolution task; it streams counterfactual facts in which later facts silently override earlier values for the same key. Streams span 455--18,332 facts (6k--262k tokens) with 55--63\% of parsed keys overwritten, and pose 100 single-hop and 100 multi-hop questions requiring current values; the three lengths are separate instances, not one truncated stream.

\vspace{0.02in}\noindent\textbf{Metrics.}
We report \emph{answer accuracy} and, on superseded facts, the \emph{stale-answer rate}---the fraction of superseded-fact questions answered with the outdated value---together with \emph{read cost} in context tokens per query (\S\ref{sec:token-efficiency}). LongMemEval accuracy and stale rate are scored by a 35B gold-aware judge. FactConsolidation uses substring-match accuracy and reports the single-hop stale rate over identifiable update chains ($n{=}64$ at 6k, $71$ at 262k); with 100 questions per cell, 95\% binomial intervals span roughly $\pm$10 points. The large multi-hop gaps at 6k and 64k (e.g.\ 41 vs.\ 14) exceed this margin; the single-hop gaps and the 262k multi-hop gap (11 vs.\ 8) fall within it, and we read those as ties or directional rather than established.

\vspace{0.02in}\noindent\textbf{Baselines and protocol.}
Within each benchmark all arms ingest identical streams and share one Qwen3-4B agent, the Qwen3-4B-Embedding embedder, and the same prompts and retrieval budget, so differences isolate the memory design rather than the backbone. Updates arrive in order and currency follows last-write-wins, so we evaluate precisely \emph{currency under arrival-ordered revisions}, not the broader valid-time setting---delayed or backdated reports, out-of-order ingestion, conflicting source authority---which the typed operations can express but which we leave to future work (Appendix~\ref{sec:limitations}).
The baselines span the corners of Table~\ref{tab:design-space}: \emph{Long-context} (most recent facts that fit the 4B window, no retrieval); \emph{MemoryOS}~\citep{kang2025memoryos} (top-five content-keyed chunk retrieval---this row doubles as the \emph{MemRL}~\citep{zhang2026memrl} baseline, whose outcome-driven reranking gets no reward signal when each question is posed once); \emph{Mem0}~\citep{chhikara2025mem0} (the CRUD corner: per-fact add/update/delete, so displaced values are overwritten); \emph{HippoRAG-v2}~\citep{gutierrez2025hipporag2} (graph-RAG over an open-domain knowledge graph, conflicts resolved at query time); \emph{Graphiti/Zep}~\citep{rasmussen2025zep} (bitemporal knowledge graph with validity intervals---updates invalidate rather than delete, conflicts arbitrated by timestamp; same 4B for extraction and retrieval, with a 35B-extractor check in Appendix~\ref{app:mabcr-extra}); and \emph{Raw facts} (retains all key--value versions, top-12 embedding retrieval), with a \emph{$+$ order tags} variant that attaches arrival indices.
\emph{P-Mem} folds each fact into a typed state cell, using arrival order for supersession and soft identity for key variants, and reads the same graph through two paths: turn-level retrieval with graph-based supersession pruning (LongMemEval), and a materialized-cell read that scans 12 cells with up to three iterative lookups (FactConsolidation, ALFWorld-Revision). On ALFWorld-Revision the \emph{commit} path applies an \textsc{accept} at write time---making the new value the head of the cell and retaining the displaced one---and queries read the cell, while the \emph{raw} path is the retrieval-only (RAG) read shared with value-aware methods such as MemRL, whose reranking has no signal to learn from here. Full configurations, seeds, and scoring protocols are in Appendix~\ref{app:experiment-configs}.

\vspace{0.02in}\noindent\textbf{What each benchmark isolates.} P-Mem reads the same provenance graph through two paths matched to the input. On conversational LongMemEval the headline read is turn-level retrieval with graph-guided supersession pruning (RQ1a); we also apply the write-time materialized frontier to the same histories (RQ1b), which currently underperforms because passing mentions escape extraction and free-form keys fragment. The materialized frontier is the primary policy on the keyed streams, ALFWorld-Revision (RQ3--RQ5) and MemoryAgentBench (RQ6). Because natural streams are almost all \textsc{accept}, we attribute the benchmark accuracy and currency gains to write-time versioned state with retention, and isolate the operations beyond \textsc{accept}---reversal (RQ4) and conflict (RQ5)---in controlled probes; we do not claim the full node/edge vocabulary is required for the accuracy numbers.

\subsection{LongMemEval: knowledge updates and read cost (RQ1--RQ2)}
\label{sec:longmemeval-results}

\begin{table}[t]
  \centering
  \small
  \caption{Knowledge-update subset: accuracy and stale-answer rate vs.\ MemoryOS retrieval on both haystack regimes}
  \label{tab:ku-ablation}
  \begin{tabular}{lcccc}
  \toprule
   & \multicolumn{2}{c}{Accuracy $\uparrow$}  & \multicolumn{2}{c}{Stale rate $\downarrow$} \\
  \cmidrule(lr){2-3}\cmidrule(lr){4-5}
   & oracle & \_s & oracle & \_s \\
  \midrule
  MemoryOS & 66.7 & 60.3 & 22.6 & 19.4 \\
  P-Mem, turn read            & 75.6 & \textbf{69.2} & 19.6 & 23.6 \\
  P-Mem, turn read $+$ prune & \textbf{76.9} & 65.4 & \textbf{5.4} & \textbf{10.9} \\
  \bottomrule
  \end{tabular}
\end{table}

P-Mem with the provenance graph improves over MemoryOS retrieval on both metrics in both regimes: accuracy by +10.2\,pp on oracle (66.7$\rightarrow$76.9) and +5.1\,pp on LongMemEval-s (60.3$\rightarrow$65.4), and the stale-answer rate from 22.6\% to 5.4\% on oracle and from 19.4\% to 10.9\% on LongMemEval-s. The two components behind this gain act on different axes.

\vspace{0.02in}\noindent\textbf{Provenance pruning (validity).} Pruning superseded turns via the graph lowers the stale rate from 19.6\% to 5.4\% (oracle) and 23.6\% to 10.9\% (\_s), and on oracle also raises accuracy (75.6$\rightarrow$76.9). On the larger \_s haystack more distractors are retrieved and pruning removes some still-needed evidence, trading 3.8\,pp of accuracy (69.2$\rightarrow$65.4) for the stale reduction---still above MemoryOS. At fixed retrieved context the graph's recorded supersession is what separates a current value from a superseded one (the pair to compare is selected at query time), so the stale rate isolates its effect. Though $n{=}78$ ($\pm$10-point intervals), the effect is paired and one-sided---eight fixed, none broken (McNemar $8$--$0$, bootstrap $p{<}0.001$)---and recurs on the superseded class ($n{=}75$), FactConsolidation ($n{=}64$--$71$), and retraction ($n{=}140$). Pruning is class-targeted, so deployment needs a gate: a router deciding prune-vs-keep from the question text alone holds overall accuracy at 51.8\% (vs 52.6\% ungated), leaves aggregation untouched, and cuts the superseded stale rate 27.4\%$\to$17.7\% (class-oracle 12.9\%; Appendix~\ref{app:prune-gate}).


\vspace{0.02in}\noindent\textbf{Full benchmark.} On all 500 questions, top accuracy is close and does not separate the methods: P-Mem's turn read reaches 52.6\% and the strongest baseline HippoRAG-v2 50.8\%, within 1.8 points, so a strong retriever recovers most of the accuracy---P-Mem's advantage is the stale rate on updates (Table~\ref{tab:ku-ablation}) and write-time behavior on MemoryAgentBench and ALFWorld-Revision. Against MemoryOS retrieval (45.8\%) P-Mem gains +6.8\,pp and leads overall, superseded, and single (Table~\ref{tab:full500}). Mem0 leads aggregation (35.9\% vs 29.3\%); no-memory scores 4.0\%, so the task requires stored history. Paired bootstrap: P-Mem$-$MemoryOS $=+6.8$ (95\% CI $[+2.4,+11.2]$, $p{=}0.001$); P-Mem$-$HippoRAG-v2 $=+1.8$ ($[-2.8,+6.4]$, n.s.).

\begin{table}[t]
  \centering
  \small
  \caption{All 500 LongMemEval-s questions, accuracy by susceptibility class.}
  \label{tab:full500}
  \begin{tabular}{lcccc}
  \toprule
   & Overall & superseded & aggregation & single \\
   & (500) & (75) & (92) & (333) \\
  \midrule
  No memory          & 4.0  & 2.7  & 0.0           & 5.4 \\
  Long-context & 27.0 & 40.0 & 5.4 & 30.0 \\
  Mem0               & 39.4 & 37.3 & \textbf{35.9} & 40.8 \\
  MemoryOS& 45.8 & 53.3 & 19.6          & 51.4 \\
  HippoRAG-v2 & 50.8 & 58.7 & 33.7 & 53.8 \\
  P-Mem (turn read)    & \textbf{52.6} & \textbf{62.7} & 29.3 & \textbf{56.8} \\
  \bottomrule
  \end{tabular}
\end{table}

The MemoryOS baseline retrieves by content; under its default key (a generic per-session descriptor) retrieval is content-blind on a 48-session haystack and accuracy falls to 11.2\% (an indexing artifact), so we report its strongest content-keyed variant. This row doubles as the MemRL baseline~\citep{zhang2026memrl}: MemoryOS is MemRL's substrate, and with each question posed once no reward reaches its reranking, so the trained method coincides with this arm (Appendix~\ref{app:config-lme}). Pruning does not carry to the full benchmark: aggregation questions (``how many films have I watched'') resemble repeated updates and are pruned as superseded, dropping overall accuracy to 45.6\% and aggregation to 15.2\%; we therefore prune only on the knowledge-update subset.

\vspace{0.02in}\noindent\textbf{Commit timing (RQ1b).} Supersession can also be committed at write time, answering from materialized state alone: the policy evaluated on FactConsolidation and ALFWorld-Revision. On naturally phrased histories it currently loses to read-time pruning. The 4B writer reaches 52.6\% accuracy at a 33.3\% stale rate, behind even the raw turn read on both axes, because updates phrased in passing escape extraction. A 35B extractor restores coverage (61.5\%, stale 20.4\%), but version chains still barely form (20{,}141 facts across 18{,}050 cells), so displaced values survive under variant keys; forcing merges (threshold $0.85\to0.70$) cuts the stale rate to 13.0\% at 3.8\,pp of accuracy. No configuration dominates the read-time prune on both axes: extraction coverage scales with the writer, key canonicalization does not (Appendix~\ref{app:writetime}).

\subsubsection{Token efficiency (RQ2)}
\label{sec:token-efficiency}
P-Mem reads a smaller context than MemoryOS retrieval at the configuration that produces its accuracy. Table~\ref{tab:token} reports mean context per query, accuracy, and accuracy per 1k tokens on LongMemEval-s.

\begin{table}[t]
  \centering
  \small
  \caption{Token efficiency on LongMemEval-s (mean context tokens per query). P-Mem reads $\sim$4$\times$ fewer tokens than MemoryOS retrieval and yields $\sim$4$\times$ more accuracy per token.}
  \label{tab:token}
  \setlength{\tabcolsep}{4pt}
  \begin{tabular}{lccc}
  \toprule
   & Context tokens $\downarrow$ & Accuracy $\uparrow$ & Acc.\ per 1k tok $\uparrow$ \\
  \midrule
  Long-context & $\sim$106k & 27.0 & 0.25 \\
  MemoryOS & $\sim$11k & 45.8 & 4.2 \\
  P-Mem (turn read)     & $\sim$3k  & \textbf{52.6} & \textbf{17.5} \\
  \bottomrule
  \end{tabular}
\end{table}

\vspace{0.02in}\noindent\textbf{Fewer tokens, higher accuracy.} P-Mem reads $\sim$3k tokens per query against MemoryOS retrieval's $\sim$11k while scoring higher (52.6 vs 45.8)---$\sim$4$\times$ more accuracy per token---from retrieval granularity (twelve turns vs five sessions), not the graph, whose contribution is validity (\S\ref{sec:longmemeval-results}), not compression. A variant that compiles memory into a materialized state (one value per key, full-state read) reaches MemoryOS-comparable accuracy at $\sim$1k tokens, an $\sim$11$\times$ reduction (Appendix~\ref{app:writetime}). Both retrieval reads stay flat as the haystack grows from 8 to 48 sessions (MemoryOS $\sim$11k, P-Mem $\sim$3k), while long-context grows linearly to $\sim$106k.

\vspace{0.02in}\noindent\textbf{Long context is not a substitute.} The full $\sim$106k-token history drops accuracy to 27.0---below every memory arm (Table~\ref{tab:full500}), aggregation at 5.4---though every prompt fits the 262k window, so the failure is comprehension, not access (0.25 accuracy points per 1k vs P-Mem's 17.5). Under a fixed budget the gap widens where budgets are tightest: at a 1k budget P-Mem already scores 50.6, above MemoryOS retrieval at its full 11k (49.2), which itself falls to 27.8 under the same cap. These are read costs; Appendix~\ref{app:cost} accounts for the write side of the materialized path, including extraction calls and tokens, storage, and the number of reads at which write-time extraction pays for itself.

\subsection{ALFWorld-Revision: write-time commit vs.\ query-time reading (RQ3--RQ5)}
\label{sec:revision-results}

LongMemEval measures staleness on isolated updates. We test when revisions \emph{accumulate}, where a query-time read reassembles the order of events instead of reading a committed value.

\subsubsection{Consolidating revisions (RQ3)}

\vspace{0.02in}\noindent\textbf{A single revision does not need consolidation.} When a fact is revised only once, the raw path is more accurate than commit. The two paths read the same update sentence and differ only in what they do with it: commit converts it into a patch, raw answers from it directly. When the update states the new value directly, both are at ceiling (100\%); when the value must be inferred from a two-step description, raw reaches 96.4\% vs.\ 92.9\% for commit, and a query-time reconstruction call trails at 87.9\% ($n{=}140$). The gap is a format cost---a structured patch loses answers freeform reading keeps. On an isolated fact, materialized state buys the compact read (30 vs.\ 591 characters at $L{=}1$, Table~\ref{tab:chain}), not accuracy.

\vspace{0.02in}\noindent\textbf{Revision chains reverse the picture.} The two paths differ structurally on a \emph{sequence} of revisions. Commit consumes events one at a time in arrival order, routing each to the queried object's cell by key, so a distractor event updates its own cell and leaves the target unchanged; the raw path must retrieve the chain at query time and reassemble event order from similarity-ranked evidence. Routing is not the bottleneck: a commit variant ingesting the full interleaved pool rather than the target chain still reaches 100\% at every $L$ (Appendix~\ref{app:config-alfrev}). We sweep $L \in \{1,2,4,8\}$ revisions per fact, interleave each target chain with four distractor chains in one pool, and retrieve the top 10 events by embedding similarity (Table~\ref{tab:chain}). Commit is flat at 100\% for all $L$; the raw path collapses: 100/72.9/42.9/25.0\% at $L{=}1/2/4/8$.

\begin{table}[t]
  \centering
  \small
  \setlength{\tabcolsep}{4pt}
  \caption{Current-value accuracy as revisions accumulate ($n{=}140$ per $L$, 5 interleaved objects, top-10). Commit stays flat; the raw read must reassemble order and collapses; reconstruction and $[t{=}i]$ tags recover part; the arrival-order oracle recovers 94--100\%.}
  \label{tab:chain}
\begin{tabular}{lcccccc}
  \toprule
   & \multicolumn{5}{c}{Accuracy (\%)} & Context (chars) \\
  \cmidrule(lr){2-6}
  $L$ & Commit & Raw (top-10) & Raw $+$ recon. & Raw $+$ $[t]$ tags & Raw (oracle order) & commit / raw \\
  \midrule
  1 & 100.0 & 100.0 & 100.0 & 100.0 & 100.0 & 30 / 591 \\
  2 & 100.0 & 72.9  & 94.3  & 97.9  & 100.0 & 30 / 621 \\
  4 & 100.0 & 42.9  & 59.3  & 80.0  & 98.6  & 30 / 640 \\
  8 & 100.0 & \textbf{25.0} & 40.0 & 64.3 & 94.3 & 30 / 641 \\
  \bottomrule
  \end{tabular}
\end{table}

\vspace{0.02in}\noindent\textbf{The failure is order reassembly, not retrieval.} 
Three controls isolate the mechanism. Retrieval is not the bottleneck: top-10 coverage of the target chain stays near 100\% across $L$, so the evidence is in context when the raw path fails. Nor is the evidence insufficient: an oracle presenting the same chain in arrival order answers at 94--100\%. What breaks is reconstructing arrival order from similarity-ranked text---at $L{=}8$ the raw errors are intermediate hops or distractor locations. Stamping each memory with its arrival index ($[t{=}i]$) and letting the reader sort helps but does not rescue it: tagged accuracy still decays 97.9\%$\to$64.3\% ($L{=}2\to8$), 35.7 points below commit (Table~\ref{tab:chain}). A query-time reconstruction call recovers part of the loss but not the trend (100/94.3/59.3/40.0\% at $L{=}1/2/4/8$), still 60 points below commit at $L{=}8$: it must still recover arrival order from the evidence. Commit never faces this cost---arrival order is free at write time.

\subsubsection{Reversibility under retraction (RQ4)}
\label{sec:retraction}

Retention's design-space claim (Table~\ref{tab:design-space}) is that it turns a committed error from permanent to transient. On the chain setup ($n{=}140$), an erroneous update moves $v_1$ to $v_2$ and, after $d$ distractors, a retraction names $v_2$ as wrong without restating $v_1$, so recovery requires $v_1$ to still exist. \emph{Retention decides reversibility}: with retained predecessors, rollback recovers $v_1$ in 100\% of committed cases at both distances, versus 0\% with retention off (the cell is deleted, not demoted). Archival raw reading recovers 97.9\% adjacent but 92.0\% at $d{=}32$ (the reassembly cost of Table~\ref{tab:chain}), and Graphiti retains $v_1$ as an invalidated edge yet recovers 0\% because no operation reopens it; even Mem0's rare recoveries come from copies it happened to retain. Full setup, per-arm outcomes (Table~\ref{tab:retraction}), and Graphiti's collapse under distance are in Appendix~\ref{app:retraction-detail}.

\begin{table}[!htbp]
  \centering
  \small
  \setlength{\tabcolsep}{4pt}
  \caption{MemoryAgentBench FactConsolidation at three history lengths ($n{=}100$ per cell; stale rate on identifiable chains, $n{=}64$--$71$). Dashes: long-context stale omitted where accuracy collapsed.}
  \label{tab:mabcr}
  \begin{tabular}{lccccccccc}
  \toprule
   & \multicolumn{3}{c}{Single-hop acc.\ $\uparrow$} & \multicolumn{3}{c}{Multi-hop acc.\ $\uparrow$} & \multicolumn{3}{c}{Stale rate (sh) $\downarrow$} \\
  \cmidrule(lr){2-4}\cmidrule(lr){5-7}\cmidrule(lr){8-10}
   & 6k & 64k & 262k & 6k & 64k & 262k & 6k & 64k & 262k \\
  \midrule
  Long-context (full stream)  & 32\% & 31\% & 0\%  & 2\% & 6\% & 0\% & 79.7\% & --    & --    \\
  MemoryOS retrieval (chunks) & 41\% & 28\% & 16\% & 8\% & 2\% & 0\% & 39.1\% & 15.9\% & 16.9\% \\
  Mem0 (CRUD memory)          & 36\% & 45\% & 30\% & 5\% & 4\% & 2\% & 43.8\% & 46.4\% & 59.2\% \\
  HippoRAG-v2 (graph RAG)     & 69\% & 47\% & 39\% & 11\% & 3\% & 6\% & 35.9\% & 24.6\% & 22.5\% \\
  Graphiti/Zep (bitemporal KG) & 15\% & 32\% & 12\% & 1\% & 1\% & 4\% & 18.8\% & 10.1\% & 15.5\% \\
  Graphiti/Zep (35B extractor) & 77\% & 86\% & -- & 18\% & 12\% & -- & 10.9\% & 10.1\% & -- \\
  Raw facts (top-12)          & 71\% & 72\% & 86\% & 17\% & 12\% & 9\% & 37.5\% & 30.4\% & 11.3\% \\
  Raw facts $+$ order tags    & 94\% & 94\% & \textbf{90\%} & 14\% & 10\% & 8\% & 6.3\% & 5.8\% & 5.6\% \\
  P-Mem (state read)            & 95\% & \textbf{95\%} & 88\% & 41\% & 21\% & 11\% & 4.7\% & 5.8\% & \textbf{4.2\%} \\
  P-Mem $+$ iterative readout   & \textbf{96\%} & 94\% & 87\% & \textbf{53\%} & \textbf{37\%} & \textbf{29\%} & \textbf{3.1\%} & \textbf{4.3\%} & 7.0\% \\
  \bottomrule
  \end{tabular}
\end{table}

\subsubsection{Concurrent conflicts: contest and resolve (RQ5)}
\label{sec:conflict}

The retraction test covers errors; a second probe covers genuine conflicts. Two reports assert different locations for one object with no recency or authority cue, eight distractors follow, and a later check settles it (counterbalanced; $n{=}140$). Before the settling event the desired behavior is to surface the conflict, not pick a side. The contest path records both candidates every time (\textsc{contest} 100\%) and flags the conflict in 100\% of pre-resolution queries; when the settling event arrives \textsc{resolve} lands every time and post-resolution accuracy is 97.9\%. A last-write-wins store never represents the conflict---one confident value, right only by coin flip. The raw store notices it (96.4\%) but cannot settle it: post-resolution accuracy is 57.1\% vs.\ 97.9\% for \textsc{resolve}. Graphiti arbitrates the conflict away on arrival---both reports stay valid in only 10.7\% of graphs, pre-resolution answers match eventual gold in 50.0\% of cases, post-resolution accuracy 39.3\%. Query-time reading re-litigates every query; a committed resolution is decided once.

\subsection{MemoryAgentBench FactConsolidation (RQ6)}
\label{sec:mabcr-results}

The revision study is controlled and single-object. MemoryAgentBench scales the same currency question to streams of hundreds to tens of thousands of counterfactual facts, and to a full slate of published memory systems.

\vspace{0.02in}\noindent\textbf{Long context alone does not track currency.} The full stream in context, no retrieval, is the weakest arm at every scale (Table~\ref{tab:mabcr}): even at 6k, where it fits, single-hop is 32\% and 79.7\% of superseded questions return the old value. Past the 16k window it sees only its most recent fraction ($\sim$20\% at 64k, 5\% at 262k) and single-hop falls to 0\% at 262k---a growing history cannot be read by a fixed window, and reading it whole does not resolve conflicts.

\vspace{0.02in}\noindent\textbf{Chunk episodes do not survive a growing history; fact-granular reads do.} As history grows $40\times$, MemoryOS single-hop falls 41\%$\to$16\% and multi-hop to 0\%, while structured-memory methods stay far more accurate (Table~\ref{tab:mabcr}) reading only $\sim$0.6k characters per query vs 8k. HippoRAG-v2, the strongest archival single-hop retriever (69\% at 6k), resolves conflicts only at query time, so it falls to 47\%/39\% (64k/262k), multi-hop $\le$11\%, stale 22--36\%---above the write-time methods.

\vspace{0.02in}\noindent\textbf{CRUD commitment gives the worst currency.} Mem0 commits at write time but overwrites, and its stale rate rises 43.8\%$\to$59.2\% (6k$\to$262k)---the only method defined at every scale that worsens with history; when its update misfires, previous and current values can no longer be separated, and single-hop stays 30--45\%, below the flat store that keeps every version. This is the design space's prediction for commitment without retention (Table~\ref{tab:design-space}): P-Mem also commits but demotes rather than deletes, holding stale at 4--7\%. Mem0 commits the change in only 8.6\% of these overrides (\S\ref{sec:retraction}), so most never overwrite; targeting conversational memory, it marks the CRUD corner, not its intended use.

\vspace{0.02in}\noindent\textbf{Single-hop currency depends primarily on order metadata.} The unordered flat store gets 71--86\% (stale 11--38\%); arrival tags raise it to 90--94\%, so retrieval usually surfaces the current value but lacks the recency signal to identify it. At equal read size the order-tagged store matches P-Mem state reads (90--94\% vs.\ 88--95\%), consolidation adding only a modest further stale reduction.

\vspace{0.02in}\noindent\textbf{Multi-hop reasoning favors write-time consolidation.} The order-tagged flat store must re-identify the current version at each hop, so errors compound: 90--94\% single-hop collapses to 8--14\% multi-hop. The gain is the representation, not a heavier reader: at a \emph{matched} single read (top-12) P-Mem's state read already separates---41\% vs.\ 14\% at 6k, 21\% vs.\ 10\% at 64k---because resolving each key once at write time gives every hop a single current value. Iterative readout is an \emph{additive} lift: it raises P-Mem to 53/37/29\% from single-shot 41/21/11\% (the 262k edge, 11\% vs.\ 8\%, is within noise, so the claim rests on 6k and 64k). A template-perfect oracle answers only 31--39\% single-shot at 6k, near P-Mem's 41\%, so the single-shot ceiling is the reader's chaining, not the write path.

Appendix~\ref{app:mabcr-extra} adds write-path and scale checks---soft-identity and reader/writer-scale ablations, a 35B-extractor Graphiti rerun (single-hop 15\%$\to$77\%, still trailing P-Mem at $8.5\times$ ingest cost), and an audit of all 18,306 writes at 262k where only 2 of 29 single-hop errors trace to the write path.

\section{Conclusion}

Long-horizon agents need memory that answers what currently holds---not content but its provenance and status. We instantiate \textsc{Memory of Memory} as \textsc{Provenant Memory}: typed operations maintain state cells over a provenance ledger. Across three benchmarks it lowers stale answers, keeps revision chains current where query-time reads collapse, and consolidates conflicts flat and CRUD stores mishandle. On natural histories materialization still trails read-time pruning---extraction and key canonicalization, not the state mechanism, are the bottleneck---so retrieval stays essential; on keyed streams, provenance-maintained current state is the first-class readout.

\bibliographystyle{plainnat}
\bibliography{references}


\appendix
\section{Implementation Details}
\label{app:implementation}

This appendix records how the provenance graph and its typed operations (\S\ref{sec:pmem}), the annotation that drives them (\S\ref{sec:memory-updates}), and materialization (\S\ref{sec:reading}) are implemented, followed by per-benchmark configurations. As noted in \S\ref{sec:reading}, mechanisms none of the reported runs exercise---usage edges, support-count ranking, learned retrieval and retention scores, and budget eviction---are part of the data model but are not activated.

\subsection{Provenance Graph}
\label{app:graph}

The graph uses four node types---evidence (observation, source, type, timestamp), version (key, value, status, validity interval), operation, and tombstone---and typed edges \emph{supported-by}, \emph{supersedes}, \emph{alternative-of}, and \emph{revokes} (\S\ref{sec:pmem}); usage edges additionally link versions to the tasks that read them. A version's status is one of \{\textsc{Active}, \textsc{Challenged}, \textsc{Alternative}, \textsc{Revised}, \textsc{Superseded}, \textsc{Revoked}, \textsc{ResolvedOut}\}; the head is the most recent \textsc{Active}/\textsc{Challenged} version with an open interval, and a functional-key invariant demotes any other open \textsc{Active} version to \textsc{Alternative}.

\subsection{Annotation Application}
\label{app:annotation}

The annotator is the frozen 4B model at temperature~0, emitting schema-constrained JSON; deterministic post-processing strips code fences, parses with a fallback, normalizes keys to lower-case underscore form, and validates enumerated fields, logging failures as \textsc{noop}. Table~\ref{tab:annotation-lowering} gives how the six annotation types lower onto edges and statuses, and how \textsc{revoke}/\textsc{reject} roll back along the \emph{supersedes} chain.

\subsection{Materialization and Readout}
\label{app:materialization}

\(\operatorname{Materialize}\) produces one cell per key: the head's value, with closed-interval versions as history, \textsc{Alternative} versions as alternatives, and support counts from \emph{supports} edges (rendering and disclosure modes, \S\ref{sec:reading}). Retrieval ranks by embedding cosine over \texttt{key: value} strings at a per-benchmark top-\(K\) cutoff; reported runs use the value-first rendering with fixed disclosure defaults, so no learned readout or retention scoring is active. On LongMemEval the headline path retrieves raw turns, not cells; cells appear only in the compiled-state variant (\S\ref{sec:token-efficiency}).

\subsection{Fact Extraction and Key Identity}
\label{app:extraction}

The extractor prompt asks the 4B model for a JSON list of \{\texttt{key}, \texttt{value}\} objects. Keys must be value-independent snake-case identifiers (\texttt{personal\_best\_5k\_time}, not \texttt{personal\_best\_27\_12}), the same attribute always yielding the same key. Values are copied verbatim, order preserved, nothing deduplicated. On LongMemEval each fact carries an assertion status in \{\texttt{asserted}, \texttt{hedged}, \texttt{reported}, \texttt{withdrawn}\} (invalid\,$\to$\,\texttt{asserted}); empty-key/value facts are dropped. Soft-identity merging is described in \S\ref{sec:memory-updates}.

\section{Experiment Configurations}
\label{app:experiment-configs}

\paragraph{Shared setup.}
All three benchmarks use Qwen3-4B-Instruct-2507 as the agent model, at temperature~0 with thinking disabled, for answering, fact extraction, pruning judgments, patch prediction, and identity confirmation. Retrieval uses Qwen3-Embedding-4B (2{,}560-dimensional). Both are served through OpenAI-compatible vLLM endpoints. LongMemEval additionally uses Qwen3.6-35B-A3B-FP8 as the gold-aware judge. Within each benchmark, compared arms share the agent, embedder, prompts, and retrieval budget.

\subsection{LongMemEval}
\label{app:config-lme}

\paragraph{Data.}
LongMemEval-s holds 500 questions with a mean of 47.7 sessions per question (range 38--62); the oracle haystack keeps only the evidence sessions, a mean of 1.9 per question (range 1--6). The knowledge-update subset is the 78 questions typed \texttt{knowledge-update}.

\paragraph{Read paths.}
P-Mem flattens every session into (date, message) turns, embeds the question and each turn, and takes the cosine top-12, rendered as dated lines. MemoryOS retrieval stores each session under a key and returns the top-5 by question-key similarity; the content-keyed variant uses the session text as the key, while the default generic descriptor ``Conversation session on \{date\}'' carries no content---the indexing choice behind the 11.2\% figure (\S\ref{sec:longmemeval-results}).

\paragraph{Supersession pruning.}
On the knowledge-update subset, supersession between retrieved turns is decided at read time: every pair with mutual cosine $\geq$0.55 and different dates is tested with a constrained binary query---do the two state different values of the queried attribute?---and on \textsc{yes} the earlier-dated turn is dropped. The date fixes the direction; the model never judges which value is older.

\paragraph{Question classes.}
The 35B judge assigns each of the 500 questions to superseded (an earlier value was later replaced), aggregation (the answer combines several values, none replacing another), or single (otherwise), given the question, gold answer, and time-stamped answer-bearing turns. All runs read the same stored classification.

\paragraph{Judging.}
Accuracy is a gold-aware yes/no match. For the stale rate the judge extracts the superseded and current values from the evidence; an incorrect prediction matching the superseded value counts as stale, and questions where the two coincide are excluded.

\paragraph{Token accounting and budget sweep.}
Context sizes are token counts under the agent model's tokenizer, averaged per query. The budget sweep (\S\ref{sec:token-efficiency}) uses budgets \{1k, 2k, 3k, 5k, 8k, 11k\}: each arm retrieves once per question, the retrieved context is truncated to the budget, and the same answerer and judge score every (arm, budget) cell.

\paragraph{Baselines.}
Mem0 (\texttt{mem0ai} 1.0.1) uses the same 4B model and embedder over a local vector store; sessions ingested in order, Mem0's LLM deciding add/update/delete per fact, top-12 retrieval. No-memory answers from empty context. Long-context places the full haystack oldest-first in the window (131k-token limit).

\subsection{ALFWorld-Revision}
\label{app:config-alfrev}

\paragraph{Facts and update events.}
Object-location facts come from 140 valid-seen ALFWorld games, one record per game keyed \texttt{\{object\}.location}. Stale records and update events are templated: ``Earlier observation: \{obj\} was in \{loc\}.'' and ``Update: \{obj\} has been moved from \{a\} to \{b\}.'' The location and object vocabulary for chain construction is harvested from the same split.

\paragraph{Single-revision variant.}
The single-revision comparison ($n{=}140$) uses a harder update phrasing: each templated update is rewritten into a two-step move description by a stronger model, and a rewrite is kept only if a temperature-0 verifier recovers the gold destination from it. The three arms (raw read, commit, query-time reconstruction) receive identical evidence.

\paragraph{Chain construction.}
For each case and chain length \(L\in\{1,2,4,8\}\), the target object gets a chain of \(L\) moves over distinct locations and four distractor objects each get their own \(L\)-move chain; the five chains merge into one arrival order by random draws across per-chain queues (preserving each chain's internal order). The retrieval pool holds the five stale records plus all \(5L\) update events. Generation is deterministic per case (seed 42).

\paragraph{Read paths.}
The query is ``where is the \{object\} now?''. The raw path embeds the pool, takes the cosine top-10, and lists the retrieved lines in similarity order. The commit path consumes update events in arrival order, routing each to its object's cell by key; each triggers one patch-prediction call emitting a \textsc{Revise} object (key, old/new value, status, superseded), and the query reads the resulting one-line cell \texttt{[key] value}. Table~\ref{tab:chain} feeds this arm the target chain directly; this equals processing the full interleaved pool, since per-object keying sends each distractor event to its own cell (a variant ingesting all $5L$ events reaches the same 100\% at every $L$, $n{=}140$). Mean read size is 30 characters for the cell vs 591--641 for the raw listing. The \([t{=}i]\) variant prefixes every pool line with its arrival index; the oracle-order variant presents the target's stale record and chain in arrival order, bypassing retrieval.

\paragraph{Scoring and controls.}
Answers are scored by normalized string match against the gold location. Retrieval coverage---the fraction of top-10 slots holding target-chain lines---stays $\geq$98.8\% for every \(L\), so the raw path's decay in Table~\ref{tab:chain} is not a retrieval failure.

\subsection{Reversibility under retraction (RQ4): full results}
\label{app:retraction-detail}
Each case reuses the chain setup ($n{=}140$, five interleaved objects, top-10 retrieval, same 4B agent): a location $v_1$ is stated once, an erroneous update moves it to $v_2$, and after $d$ distractor events a retraction states the move to $v_2$ was reported in error. The retraction names $v_2$ but never restates $v_1$, which appears nowhere else, so answering (gold $v_1$) requires that $v_1$ still exist in the store. The erroneous update is indistinguishable from a legitimate one at write time---all three arms commit it (pre-retraction queries answer $v_2$ in 97.9--100\% of cases)---and post-retraction outcomes are over committed cases only.

\paragraph{Retention decides reversibility.} With retention, rollback lands on the retained prior version and recovery is 100\% at both distances (Table~\ref{tab:retraction}). With retention disabled, the same rollback has nothing to return to: recovery is 0\%, every case lost, the cell deleted not restored. The mechanism check matches: $v_1$ survives in 100\% of retention stores and 0\% of CRUD ones.

\paragraph{Archival stores recover at reassembly cost.} The raw store retains $v_1$ as text and recovers 97.9\% of adjacent cases but 92.0\% at $d{=}32$: connecting the statement, the update, and the retraction from similarity-ranked evidence weakens with distance, the reassembly failure of Table~\ref{tab:chain}. Mem0 committed the error in only 8.6\% of cases, and $v_1$ survived separately in 11 of those 12, so its recoveries came from copies it happened to retain; even in a CRUD design, recovery tracks retention.

\paragraph{Retention without a rollback edge does not reverse.} Graphiti commits the adjacent error in 135 of 140 cases and retains $v_1$ in every store as an invalidated edge, yet recovers 0\% at both distances: the retraction is one more fact, and no operation re-opens an invalidated edge. At $d{=}32$ commitment itself falls to 11.4\%; under distractor load the recency-first readout buries even the committed error.

\begin{table}[t]
  \centering
  \small
  \caption{Reversibility under retraction ($n{=}140$, outcomes over committed cases only): recovered $=v_1$, wrong $=v_2$, lost $=$ neither. The two commit arms differ only in retention. Graphiti/Zep commits only 16 errors at $d{=}32$, a small base for its row there.}
  \label{tab:retraction}
  \begin{tabular}{lccccc}
  \toprule
   & \multicolumn{2}{c}{Recovered $\uparrow$} & Wrong & Lost & $v_1$ in store \\
  \cmidrule(lr){2-3}
   & $d{=}0$ & $d{=}32$ & \multicolumn{2}{c}{($d{=}32$)} & \\
  \midrule
  Commit $+$ retention & \textbf{100\%} & \textbf{100\%} & 0\% & 0\% & 100\% \\
  Commit, no retention         & 0\% & 0\% & 0\% & 100\% & 0\% \\
  Raw top-10 (archival)        & 97.9\% & 92.0\% & 6.6\% & 1.5\% & 100\% \\
  Graphiti/Zep & 0\% & 0\% & 93.8\% & 6.2\% & 100\% \\
  \bottomrule
  \end{tabular}
\end{table}

\subsection{MemoryAgentBench FactConsolidation}
\label{app:config-mabcr}

\paragraph{Data.}
We use FactConsolidation, the Conflict-Resolution split of the official release. Single-hop and multi-hop questions at each length share one stream; the three lengths hold 455, 4{,}580, and 18{,}332 numbered facts (6k, 64k, 262k tokens), 100 questions per cell. Template parsing identifies 239, 2{,}263, and 8{,}445 keys, of which 131, 1{,}277, and 5{,}321 are overwritten at least once (the 55--63\% overwrite rates in \S\ref{sec:mabcr-results}).

\paragraph{Write path.}
The stream is chunked into 1{,}000-character line-aligned pieces and the 4B model extracts \{\texttt{key}, \texttt{value}\} facts per chunk (Appendix~\ref{app:extraction}). Facts are written sequentially through the patch interface: the first write of a key creates its cell, each later write supersedes the head, so stream order determines supersession. No cap or eviction applies; the store grows to 13{,}420 cells at 262k. Soft identity merges 37 of 133 gated candidates at 6k, reducing 346 cells to 307 against 239 template keys.

\paragraph{Read path.}
Cells are embedded as \texttt{key: value} strings, retrieved by cosine top-12, and rendered one line per cell under a tracked-state header. The answer prompt states that the later of two conflicting facts is current and that memory overrides world knowledge. Iterative readout lets the model issue a lookup (naming an entity and relation) or a final answer; each lookup retrieves six further cells, deduplicated, with at most three before an answer is forced (observed mean 1.5 at 6k). MemoryOS is also the substrate of MemRL~\citep{zhang2026memrl}, which reranks by outcome-driven Q-values; with each question posed once the Q term stays at its prior, so the shipped MemRL scores 58.2\% on oracle and matches the untrained substrate on LongMemEval-s (11.2\% default key, 45.8\% content-keyed), and its MemoryOS-retrieval arm plays the same role here.

\paragraph{Baselines.}
Long-context keeps the most recent 56{,}000 characters ($\sim$13k tokens of the 16k window): all of 6k, 20.5\% at 64k, 5\% at 262k. MemoryOS retrieval stores each 1{,}000-character chunk under its own text and returns the top five. Mem0 (\texttt{mem0ai} 1.0.1) ingests the same chunks with a domain fact-extraction prompt, add/update/delete unmodified, top-12 retrieval. Raw facts keeps every version in a flat store, cosine top-12 in retrieval order; the order-tag variant sorts by stream index and prefixes it. The oracle write path regex-parses the template chains and keeps the last value per key (239 cells at 6k), read top-12 or as full state.

\paragraph{Metrics.}
Accuracy is normalized exact-or-substring match; no LLM judge. A question enters the stale-rate denominator when a gold answer equals the final value of a key with more than one parsed value ($n{=}64$, 69, 71 single-hop at the three lengths); a prediction is stale when incorrect and matching an earlier value of that key.

\section{Gating the Provenance Prune}
\label{app:prune-gate}

The Table~\ref{tab:ku-ablation} pruning is class-targeted; a deployed system must decide per query. We test gates choosing between the raw and pruned read from the question text alone: an intent classifier (prune only when the question asks for a current value) and a binary prune-vs-keep router, with the 4B and the 35B. No gate sees the class or the retrieved turns, and every arm reuses the 500-question run's stored answers and judge verdicts, so the comparison is paired.

\begin{table}[h]
  \centering
  \small
  \setlength{\tabcolsep}{4pt}
  \caption{Gating the provenance prune on all 500 LongMemEval-s questions (\%). Per-class accuracy over the judge-assigned classes; the stale rate is over the 62 stale-scoreable superseded questions of this run, a different subset from the 78-question knowledge-update set of Table~\ref{tab:ku-ablation}. Pruned = fraction of questions the gate routes to the pruned read.}
  \label{tab:prune-gate}
  \resizebox{\columnwidth}{!}{%
  \begin{tabular}{lcccccc}
  \toprule
  Gate & Overall $\uparrow$ & Agg. & Single & Superseded & Stale $\downarrow$ & Pruned \\
  \midrule
  Ungated raw read            & 52.6 & 29.3 & 56.8 & 62.7 & 27.4 & 0 \\
  Prune always                & 45.6 & 15.2 & 51.4 & 57.3 & \textbf{12.9} & 100 \\
  Class-oracle gate           & 51.8 & 29.3 & 56.8 & 57.3 & \textbf{12.9} & 15.0 \\
  Intent gate (4B)            & 51.2 & 29.3 & 55.6 & 58.7 & 25.8 & 17.2 \\
  Question-only router (4B)   & 47.8 & 22.8 & 52.0 & 60.0 & 16.1 & 68.8 \\
  Question-only router (35B)  & 51.8 & 29.3 & 55.3 & \textbf{64.0} & 17.7 & 22.2 \\
  \bottomrule
  \end{tabular}%
  }
\end{table}

Unconditional pruning collapses aggregation (29.3\% to 15.2\%) and costs 7.0 points overall, which is why the main-table P-Mem row is ungated. The 35B question-only router holds overall accuracy within 0.8 points of ungated, leaves aggregation untouched, and cuts the superseded stale rate from 27.4\% to 17.7\% (one-sided: six fixed, none broken), recovering about two thirds of the class-oracle reduction (12.9\%). Router quality matters: the 4B binary router prunes 68.8\% and drops to 47.8\% overall, and the 4B intent gate protects accuracy (51.2\%) but routes the wrong questions, leaving stale at 25.8\%. The gate is a one-token classification per query, independent of the write path.

\section{Write-Time Materialization on Natural Histories}
\label{app:writetime}

This section details the commit-timing comparison of \S\ref{sec:longmemeval-results}. The write path of \S\ref{sec:memory-updates} runs unchanged on the naturally phrased \_s knowledge-update subset: facts are extracted from each session in arrival order, supersession is committed at ingestion by arrival order on an embedding-canonicalized key (the embedding match of soft identity, without its confirmation check), and the 4B reader answers from the materialized state alone, so no commitment or retrieval decision sees the question. Both read-time baselines are re-evaluated under the same 35B judging pass as the write-time arms; they match Table~\ref{tab:ku-ablation} within one question (this pass scores 54 stale-scoreable updates against 55 there).

\begin{table}[h]
  \centering
  \small
  \caption{Write-time materialization on the \_s knowledge-update subset ($n{=}78$; stale rate over the 54 stale-scoreable updates of this judging pass). All arms share the 4B reader; write-time arms read the full materialized state. Facts/cell measures whether supersession chains form (1.0 = every fact its own cell).}
  \label{tab:writetime}
  \begin{tabular}{lccc}
  \toprule
   & Accuracy $\uparrow$ & Stale $\downarrow$ & Facts/cell \\
  \midrule
  Turn read, raw & \textbf{69.2} & 24.1 & --- \\
  Turn read + provenance prune & 65.4 & \textbf{9.3} & --- \\
  \midrule
  Materialized state, 4B writer & 52.6 & 33.3 & 1.17 \\
  \quad 35B extractor & 61.5 & 20.4 & 1.12 \\
  \quad 35B extractor, merge 0.70 & 57.7 & 13.0 & 1.76 \\
  \bottomrule
  \end{tabular}
\end{table}

Table~\ref{tab:writetime} reports every arm under this pass. The 4B writer's failures are extraction coverage, not commitment: updates delivered in passing escape the fact extractor. One question's answer changes from \$350{,}000 to \$400{,}000 inside an aside about looking forward to owning a home; the 4B extractor captures neither value, while the turn retriever surfaces the raw message. The deficit is paired and significant (McNemar $9$--$22$ against the raw read, $p{=}0.03$). The 35B extractor (same convention as the Graphiti rerun of Appendix~\ref{app:mabcr-extra}: a labeled arm, not the shared-backbone comparison) captures both and commits the update, recovering 8.9\,pp of accuracy and returning the stale rate to the raw read's level (no longer distinguishable from it, $p{=}0.42$); the fragmentation measured on template streams in Appendix~\ref{app:mabcr-extra} is amplified by natural phrasing. Fragmentation does not yield to writer scale: facts per materialized cell moves from 1.17 (4B) to 1.12 (35B), against 55--63\% of keys overwritten on FactConsolidation, so most cells hold a single fact and a displaced value survives as a sibling cell rather than a superseded predecessor. Lowering the embedding-match threshold from 0.85 to 0.70 merges aggressively (1.76 facts per cell): chains form, the stale rate falls to 13.0\%, and false merges take back 3.8\,pp of accuracy. A matched-budget read over the twelve cells nearest the question does not change the picture (56.4\%/20.4\% for the 4B writer, 56.4\%/25.9\% for the 35B, 52.6\%/16.7\% at threshold 0.70; the last is behind the raw read at $p{=}0.047$). Write-side cost for the 78 histories (3{,}706 sessions): one extraction call per session, 17.0 minutes with the 4B writer and 14.1 with the 35B on one H200; the full-state read averages 39k--84k characters against roughly 3k for the twelve-cell read. Across both writer scales and both merge thresholds the write-time frontier stays inside the read-time pruning point of Table~\ref{tab:ku-ablation} on both axes.

\section{Write-Path Cost Accounting}
\label{app:cost}

Table~\ref{tab:writecost} accounts for the write side of the materialized path on the 78 \_s knowledge-update histories, measured from per-request logs; judging is excluded as evaluation harness. Extraction is one LLM call per session (47.5 per history; 0.15--0.18 calls per committed fact, since one session yields several facts), 146--157k LLM tokens per history, and 11--13 seconds of wall time per history at 20-way request concurrency on one H200. Each extracted key costs one embedding call (51 tokens on average), and cells are embedded once more for the matched-budget read. Storage is dominated not by the extracted facts (0.46--0.62\,MB per history as JSON) but by the embedding index (2.4--2.8\,MB per history at fp32, one 2{,}560-dimensional vector per cell).

\begin{table}[h]
  \centering
  \small
  \caption{Write-side cost per history on the \_s knowledge-update subset (78 histories, 47.5 sessions each), measured from per-request logs.}
  \label{tab:writecost}
  \begin{tabular}{lcc}
  \toprule
   & 4B writer & 35B extractor \\
  \midrule
  Extraction LLM calls & 47.5 & 47.5 \\
  Calls per committed fact & 0.15 & 0.18 \\
  Write tokens (prompt$+$completion) & 157k & 146k \\
  Ingestion wall time & 13.1\,s & 10.8\,s \\
  Committed facts & 325 & 258 \\
  Materialized cells & 278 & 231 \\
  Fact store (JSON) & 0.62\,MB & 0.46\,MB \\
  Embedding index (fp32) & 2.8\,MB & 2.4\,MB \\
  \bottomrule
  \end{tabular}
\end{table}

\paragraph{Break-even.} Write-time extraction pays for itself in read savings after roughly 70 queries per history: extraction costs $\sim$157k tokens once, and a twelve-cell state read ($\sim$0.7k tokens) saves $\sim$2.3k tokens against the twelve-turn read of Table~\ref{tab:token} ($\sim$3k). Against a full-history read ($\sim$106k tokens per query) a single query repays it. The count charges nothing for the read-time path's per-query pruning verifications, which would lower the break-even further, and it prices write and read tokens equally. At LongMemEval's one question per history the write path therefore does not pay for itself economically either, consistent with Appendix~\ref{app:writetime}; a long-lived assistant that consults the same state across many sessions crosses the threshold quickly. Operation counts and the per-fact audit of the FactConsolidation write path, and Graphiti's ingestion costs under the same conditions, are in Appendix~\ref{app:mabcr-extra}.

\section{Scale and Write-Path Analyses}
\label{app:mabcr-extra}

\paragraph{Write-path ablation: soft identity.} At 6k the extractor creates 346 cells for 239 template keys, leaving superseded values active in separate cells. Soft-identity merging (\S\ref{sec:memory-updates}) cuts single-hop stale rates from 9.4\% to 4.7\% at 6k and 7.0\% to 4.2\% at 262k, and raises iterative multi-hop from 42\% to 53\% and 24\% to 29\%. At 262k the write path applies 18,306 operations, converting about 1,800 Creates into Revises.

\paragraph{A stronger reader moves single-hop, not multi-hop.} Swapping the 4B reader for a 35B at 6k (stores still built by the 4B writer) lifts the unordered flat store from 71\% to 94\% single-hop (stale 37.5\% to 3.1\%); single-shot arms converge to 94--98\%. Multi-hop barely moves (flat 21 to 23\%, P-Mem 43 to 52\%) and the iterative ceiling holds (52\% vs 53\%), so the $\sim$2$\times$ consolidation gap persists---the residual difficulty is chaining current values, not reader scale.

\paragraph{Writer scale.} Rebuilding every store with a 35B write path (the 4B remains reader) moves all arms by at most three points at 6k: the state read reaches 98\% single-hop at 0\% stale, iterative multi-hop 54\% vs 53\%, the flat store 65\% vs 71\%. The write path is not the bottleneck: a template-perfect write-path read single-shot reaches only 31--39\% on multi-hop, comparable to P-Mem's single-shot state read.

\paragraph{Reader scale on revision chains.} With the commit write path fixed at 4B, a 35B reader lifts the raw path's chain reassembly only modestly---72.9\% to 87.1\% at $L{=}2$, 25.0\% to 33.6\% at $L{=}8$---while an arrival-order oracle reads 100\% at every $L$. The reassembly failure is structural, not reader scale; commit stays at 100\%.

\paragraph{Write-path audit.} The template-generated stream makes the gold state mechanically derivable: parsing the fact list yields every key's value sequence, checked against each active cell with no LLM judging. At 262k the current value survives for 99\% of the 8,445 gold keys (84\% in a cell whose key also matches), and 13.0\% of the 5,321 overwritten keys retain a stale sibling, almost entirely revise-as-create fragmentation; soft identity halves this at every scale (22.9$\to$9.9\% at 6k, 30.3$\to$13.0\% at 262k). The imperfection barely propagates: of the main run's 29 single-hop errors across scales, only 2 trace to the write path (gold value absent from every active cell); the rest are read errors on a store that holds the answer. With 100\% commit-step success on templated and paraphrased chains and 100\% rollback and contest/resolve fidelity in the probes, this bounds the write path's contribution to end-task error.

\paragraph{Graphiti/Zep extraction on a local model.} Run on the same 4B, the bitemporal graph is limited by extraction, not conflict resolution. At 6k it builds 322 entity nodes and 110 relation edges from 455 facts over 239 keys. The gold current value appears in the top ten retrieved edges for only 32 of 100 single-hop questions, capping accuracy near there; the observed 15\% sits below because retrieved sets often hold both current and superseded values and the 4B reader does not always pick the current one. Its low stale rate (18.8\%) is an abstention artifact---a fact never retrieved cannot be stale---not conflict-resolution skill. Table~\ref{tab:mabcr} reports it as-is; it targets conversational memory, not streamed counterfactual consolidation.

\paragraph{Graphiti/Zep with a strong extractor.} Rerunning 6k with the 35B as Graphiti's internal model (extraction, deduplication, invalidation, reranking; the 4B answerer unchanged) confirms extraction was the binding constraint: relation edges grow 110$\to$367 over the same 455 facts, gold-in-top-ten recall 32\%$\to$78\%, single-hop 77\%, stale a real 10.9\%. The write-time methods still lead on all three axes---96\% vs.\ 77\% single-hop, 3.1\% vs.\ 10.9\% stale, 53\% vs.\ 18\% multi-hop, where Graphiti matches the order-tagged flat store (14\%) as each hop resolves currency at query time. Costs do not transfer: ingestion slows 15.6$\to$133 s per chunk while P-Mem's write path barely moves 4B$\to$35B. The pattern holds at 64k (recall 92\%): 86\% vs.\ 95\% single-hop, 10.1\% vs.\ 4.3--5.8\% stale, 12\% vs.\ 37\% multi-hop (12.5 h ingestion); 262k out of reach.

\section{Limitations}
\label{sec:limitations}

Our experiments exercise only part of the operation vocabulary---almost all \textsc{accept}, with \textsc{contest}/\textsc{resolve} only in a controlled probe (\S\ref{sec:conflict})---on one 4B agent (35B reader/writer checks leave the gaps intact, Appendix~\ref{app:mabcr-extra}), so natural-stream operation frequency, larger models, and other benchmarks are untested. Updates are single sentences---paraphrasing leaves every commit step intact (\S\ref{sec:revision-results}), but implied-outcome ones could introduce write-path errors---and questions follow ingestion, not repeated interaction. Two deployment gaps remain: append-only demotion is not erasure, so privacy deletion must hard-delete nodes, embeddings, and derived values; and the annotator does not weigh source authority, so an untrusted observation can supersede a trusted value---memory poisoning we leave to future work.

\end{document}